\documentclass{article}
\usepackage{iclr2027_conference,times}

\usepackage{graphicx}
\usepackage{url}
\usepackage{amsmath,amssymb}
\usepackage{booktabs}
\usepackage{xspace}
\usepackage{float}
\usepackage{xcolor}
\usepackage{hyperref}
\usepackage{pifont}
\usepackage{wrapfig}
\usepackage{xcolor}
\usepackage{colortbl}
\usepackage{algorithm}
\usepackage{algpseudocode}
\usepackage{graphicx}
\definecolor{fixedmodel}{HTML}{FD36A7}
\definecolor{mllmrouting}{HTML}{00A5FB}
\definecolor{systemone}{HTML}{4ADDB3}

\makeatletter
\DeclareRobustCommand{\onedot}{\futurelet\@let@token\@onedot}
\def\@onedot{\ifx\@let@token.\else.\null\fi\xspace}
\makeatother
 
\def\ie{\emph{i.e}\onedot}

\title{EditJev: System One Routing for Image Editing Models}

\author{
\resizebox{\textwidth}{!}{%
Zhile Guan\textsuperscript{1,*}\quad
Xu Jiang\textsuperscript{1,*}\quad
Yule Duan\textsuperscript{1,*}\quad
Gongbo Zhang\textsuperscript{1}\quad
Shuyu Wang\textsuperscript{1}\quad
Xiao Chen\textsuperscript{2}\quad
Jian Zhang\textsuperscript{1,$\dagger$}%
}\\[0.5em]
\resizebox{\textwidth}{!}{%
\textsuperscript{1}School of Electronic and Computer Engineering, Peking University
\quad
\textsuperscript{2}Tsinghua University
}
}

\date{}

\hypersetup{
  pdftitle={EditJev: System One Routing for Image Editing Models},
  pdfauthor={Zhile Guan, Xu Jiang, Yule Duan, Gongbo Zhang, Shuyu Wang, Xiao Chen, Jian Zhang}
}
\iclrfinalcopy

\begin{document}

\maketitle

\begingroup
\renewcommand{\thefootnote}{\fnsymbol{footnote}}
\footnotetext[1]{Equal contribution.
\qquad
$\dagger$ Corresponding author.}
\endgroup

\fancyhead{}
\renewcommand{\headrulewidth}{0pt}

\begin{abstract}

Instruction-based image editing models exhibit pronounced sample-specific performance variation: different models can produce substantially different results for the same source image and editing instruction, and the best-performing model often varies across requests. These models also differ substantially in inference latency, making fixed-model deployment inefficient. We therefore study image editing model routing, which dynamically selects an editor from a candidate pool given an input image and editing instruction. We introduce EditJev, a System One routing model that directly predicts sample-level preference scores over candidate image editing models in a single forward pass. A lightweight preference head on a vision-language backbone produces a fixed-dimensional score vector without autoregressive model-name generation or response parsing. To train EditJev, we construct pairwise preferences from offline candidate outputs and optimize a pairwise ranking objective to learn relative model suitability. At inference time, EditJev combines the predicted preference scores with model-level speed scores to balance editing quality and generation latency. Experiments on ImgEdit and GEdit-En show that EditJev improves aggregate quality over the strongest fixed editor, from 4.540 to 4.592 and from 8.194 to 8.255, respectively, while reducing estimated end-to-end latency by 60.8\% and 31.7\%. These results demonstrate the effectiveness of direct, structured model routing for efficient image editing.
\end{abstract}

\begin{figure}[H]
\centering
\includegraphics[width=\textwidth]{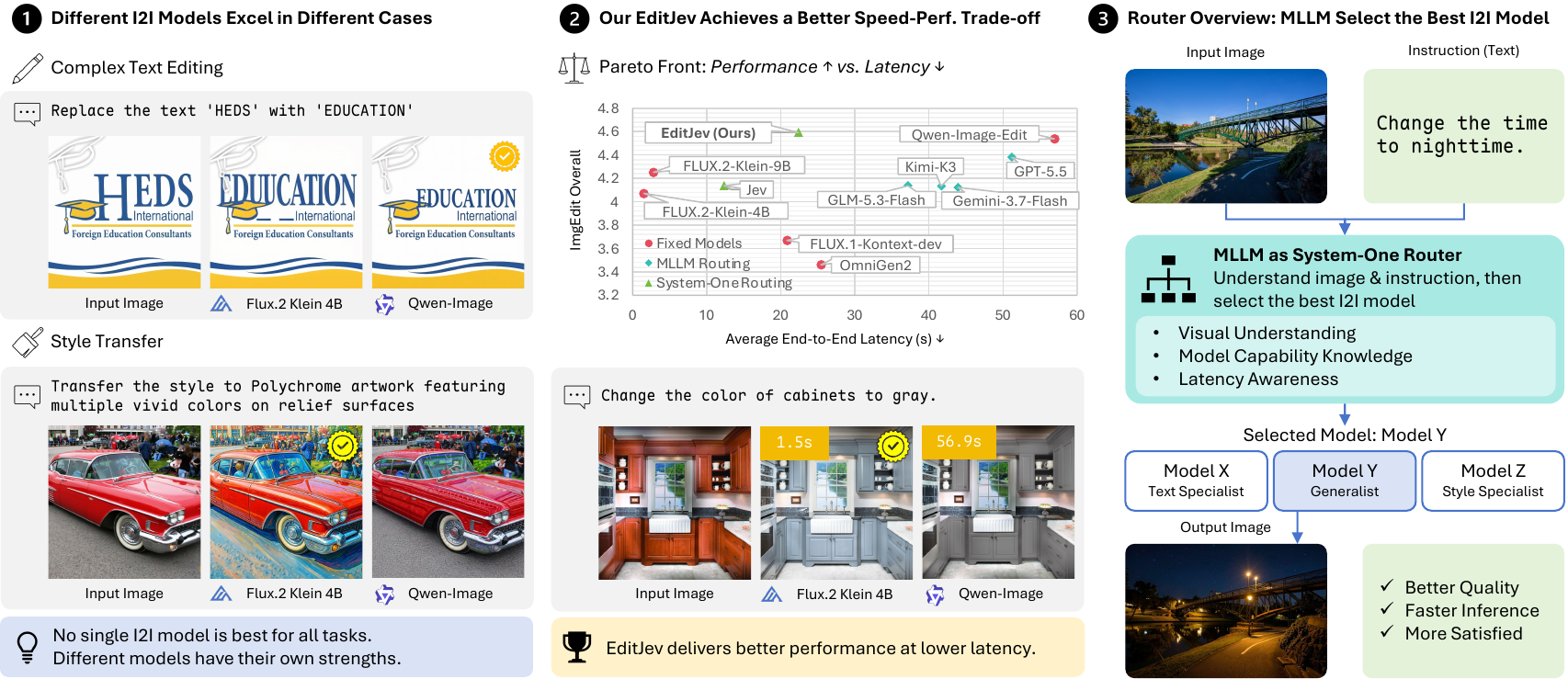}
\caption{
\textbf{Left:} Different image editing models exhibit complementary strengths: the optimal model varies across image-instruction pairs, and no single model consistently performs best for all editing requests.
\textbf{Middle:} This observation motivates model routing, where EditJev achieves a better quality--latency trade-off by dynamically selecting suitable models instead of relying on a fixed model.
\textbf{Right:} Given an input image and instruction, EditJev uses a multimodal router to understand the request, estimate model suitability, and select the model that provides better editing quality with lower inference latency.}
\label{fig:editJev-teaser}
\end{figure}

\section{Introduction}

Instruction-based image editing enables users to modify an image through natural-language commands. A single instruction can specify operations ranging from local object insertion, removal, and attribute adjustment to global style transfer and illumination changes~\citep{brooks2023instructpix2pix,zhang2024magicbrush,sheynin2024emuedit,huang2024smartedit}. Advances in editing architectures, training data, and model scale have rapidly expanded both the capability and the diversity of available models~\citep{wu2025omnigen2,flux2025kontext,qwenimageedit,hunyuanimage3}.

Yet these models are not interchangeable. Differences in architecture, training strategies, and data produce complementary behavior across editing tasks~\citep{zang2026instruction,hertz2022prompt,huang2025diffusion}. A model that performs creative global restyling may alter fine details or spatial structure, whereas a model specialized in precise local manipulation may be less effective at globally consistent transformations~\citep{chen2025edival}. Their computational profiles are similarly heterogeneous: larger and slower models can be more capable, but they also incur substantially greater memory use and inference latency~\citep{qwenimageedit,hunyuanimage3}. Thus, model quality and cost depend not only on the model itself but also on the particular image-instruction pair.

This heterogeneity creates a practical deployment dilemma. As illustrated in the left panel of Fig.~\ref{fig:editJev-teaser}, different image editing models exhibit complementary strengths across image-instruction pairs~\citep{team2026firered,ye2026imgedit}.
Sending every request to a single default model, typically the largest or most capable overall, can waste computation on easy inputs while overlooking alternatives better suited to particular requests~\citep{zhang2026router,jiang2026multi,feng2026llmrouter,frugalgpt}.
Yet manually selecting an editor for each request is impractical, as it requires substantial human effort and detailed knowledge of the candidate models. 

Motivated by this observation, we formulate the problem as image editing model routing: given an image, a natural-language instruction, and a set of candidate editors, select a suitable editor before generation while accounting for editing quality and inference cost. The central challenge is that the quality each candidate would achieve is unknown at the time of selection. A router must infer the most suitable editor directly from the input image-instruction pair.


To address this challenge, we introduce EditJev, a System One routing model that learns to predict sample-level preferences over candidate image editing models to guide editor selection. Following the System One principle of efficient decision making through structured prediction~\citep{typesafe2026introduction}, EditJev represents these preferences as a compact, fixed-dimensional score vector. As illustrated in the right panel of Fig.~\ref{fig:editJev-teaser}, EditJev equips a vision-language backbone with a lightweight preference head to produce a fixed-dimensional score vector in a single forward pass, requiring only 94 ms for routing inference.

We construct training supervision by evaluating editing outputs generated offline by all candidate models using a multimodal judge. We convert these evaluations into pairwise preferences and train EditJev with a pairwise ranking objective, enabling it to learn relative model suitability for each image–instruction pair. At inference time, as illustrated in the middle panel of Fig.~\ref{fig:editJev-teaser}, EditJev combines its predicted preference scores with model-level speed scores to select an editor according to the desired balance between editing quality and generation latency.

Our contributions are summarized as follows:

\vspace{2pt}
\noindent \ding{113}~\textbf{Problem formulation.}
To the best of our knowledge, we are the first to formulate sample-level routing for instruction-based image editing, where a router selects among candidate editing models by jointly considering editing quality and inference latency.

\vspace{2pt}
\noindent \ding{113}~\textbf{System One routing framework.}
We develop EditJev, a System One routing model that directly predicts preference scores over candidate editors in a single forward pass. Gap-filtered pairwise supervision enables the model to learn the relative suitability of candidate editors and supports latency-aware model selection at inference time.

\vspace{2pt}
\noindent \ding{113}~\textbf{Improved quality--latency trade-off.}
We evaluate EditJev on ImgEdit and GEdit-En, where it improves aggregate quality over the strongest fixed editor while reducing estimated latency, including routing overhead, by 60.8\% and 31.7\%, respectively.

\section{Related Work}

\subsection{Instruction-Based Image Editing}

Instruction-based image editing modifies an input image according to a natural-language instruction. Prior work has advanced this task through image-instruction datasets~\citep{brooks2023instructpix2pix,zhang2024magicbrush,hui2024hq,zhao2024ultraedit} and dedicated editing frameworks~\citep{sheynin2024emuedit,huang2024smartedit,li2025uniworldv2}, supporting diverse operations such as object insertion and removal, attribute modification, and style transformation~\citep{yu2025anyedit,liu2025step1x,xu2025insightedit}. Recent methods further improve editing quality by scaling model capacity, training data, and compute~\citep{wu2025omnigen2,flux2025kontext,qwenimageedit,hunyuanimage3,team2025longcat,team2026firered}. For example, Qwen-Image-Edit has 20 billion parameters, while HunyuanImage-3.0-Instruct has 80 billion parameters in total~\citep{qwenimageedit,hunyuanimage3}. Such scaling, however, increases inference latency and deployment cost. Since editing requests vary in difficulty, lightweight models may suffice for some inputs, whereas others require stronger models, motivating dynamic routing to balance editing quality and computational cost.

\subsection{Model Routing}

Model routing dynamically selects a model from a candidate pool for each input, typically balancing output quality and inference cost. It has been extensively studied for large language models(LLMs) as a strategy for cost-aware serving~\citep{hu2024routerbench,lai2026routejudge}. Existing LLM routers select models by estimating query difficulty, model capability, or expected quality-cost trade-offs~\citep{routellm,hybridllm,feng2025graphrouter}. Model routing has also been extended to text-to-image (T2I) generation~\citep{xin2025adaptive,jiang2026octot2i}. However, model routing for instruction-based image editing remains underexplored.

Recently, System One models have explored direct structured decision making, where models output compact predictions that can be consumed by downstream systems without additional parsing~\citep{typesafe2026introduction}. This design principle is particularly relevant to routing problems, where the objective is to select among a fixed candidate set rather than generate open-ended responses. In this work, we introduce EditJev, a direct preference prediction framework that extends model routing to image editing, enabling instance-level selection among editing models.

\section{Method}

\subsection{Problem Formulation}

\textbf{Setting. }Let $\mathcal{X}$ denote the space of source images and $\mathcal{C}$
denote the space of editing instructions. We consider a fixed ordered candidate list of image editing models 
$\mathcal{M}=(M_1,\ldots,M_K)$, where $K$ is the number of candidate models and $[K]=\{1,\ldots,K\}$ denotes the candidate-model index set. Given an editing request 
$(\mathbf{x},\mathbf{c})\in\mathcal{X}\times\mathcal{C}$, a routing policy 
$\pi:\mathcal{X}\times\mathcal{C}\rightarrow[K]$ selects the index of one candidate model for execution:
\begin{equation}
  \hat{i} = \pi(\mathbf{x},\mathbf{c}),
  \qquad
  \hat{M}=M_{\hat{i}}.
  \label{eq:routing-policy}
\end{equation}

For each candidate model $M_i$, let $Q_i(\mathbf{x},\mathbf{c})$ denote the final quality score of its edited result for request $(\mathbf{x},\mathbf{c})$, and let $t_i$ denote its mean end-to-end generation latency. A desirable routing policy should select a model that achieves a favorable quality--latency trade-off, \ie, high editing quality with low inference latency.

\textbf{Challenge.} At inference time, however, $Q_i(\mathbf{x},\mathbf{c})$ is unknown before candidate model $M_i$ is executed. The key challenge is to learn an efficient routing policy that selects a suitable candidate model using only the source image and editing instruction, without observing any generated outputs.

\subsection{System One Routing}

\begin{figure*}[t]
\centering
\includegraphics[width=\textwidth]{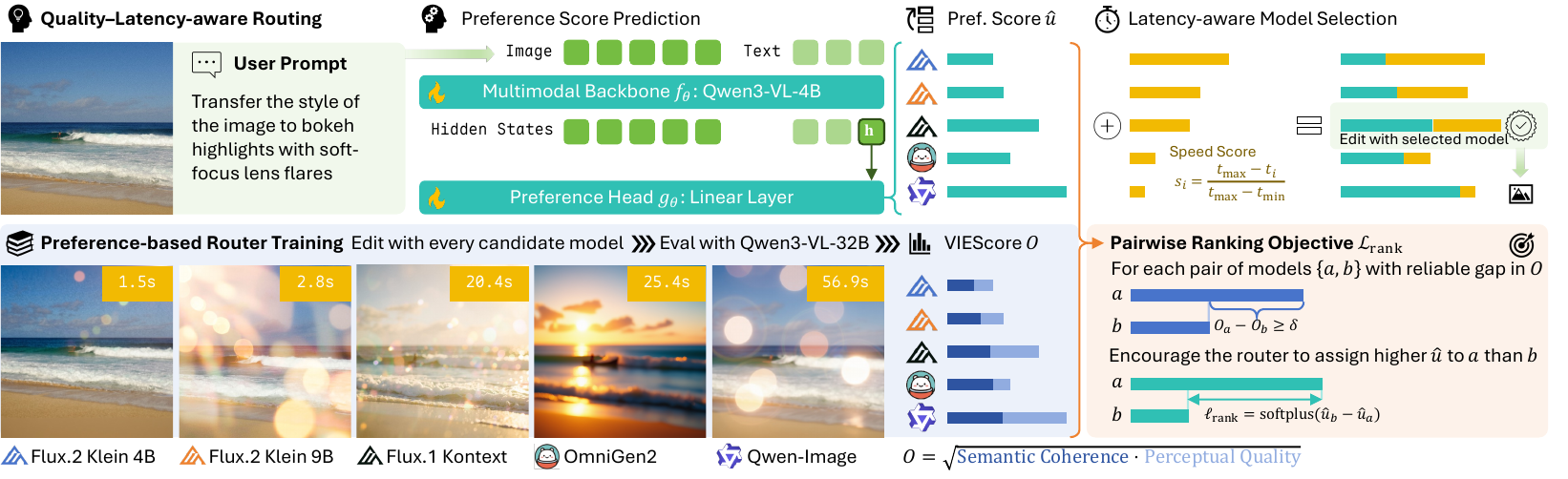}
\caption{Overview of EditJev. The upper part shows offline candidate generation, automatic evaluation, and router training. The lower part shows quality--latency-aware inference, which combines preference scores with model-level speed scores.}
\label{fig:EditJev-overview}
\end{figure*}

\textbf{Overview.} We formulate image editing routing as a System One decision problem, where the router directly predicts numerical preference scores over candidate models rather than autoregressively generating a model choice. As illustrated in Figure ~\ref{fig:EditJev-overview}, EditJev consists of offline preference-based router training and quality--latency-aware inference.
Given a source image and an editing instruction, EditJev uses a multimodal large language model (MLLM) backbone to encode the request and the candidate model set, and attaches a lightweight preference head to predict a preference score for each candidate. These preference scores are combined with model-level latency statistics to select one model for execution, enabling quality--latency-aware routing without invoking all candidate models.

\textbf{Preference Score Prediction.} Formally, given a source image $\mathbf{x}$, an editing instruction $\mathbf{c}$, and an ordered candidate model list 
$\mathcal{M}=(M_1,\ldots,M_K)$, EditJev
uses a fixed multimodal routing template to construct the prompt $R(\mathbf{x},\mathbf{c},\mathcal{M})$, which describes the editing request and lists the candidate models. The MLLM backbone $F_{\boldsymbol{\theta}}$ encodes the prompt into hidden states $\mathbf{H}\in\mathbb{R}^{L\times d}$, from which we take the hidden state of the last non-padding token as the request representation:

\begin{equation}
  \mathbf{H}
  =
  F_{\boldsymbol{\theta}}\bigl(R(\mathbf{x},\mathbf{c},\mathcal{M})\bigr),
  \qquad
  \mathbf{h}
  =
  \mathbf{H}_{j}\in\mathbb{R}^{d},
  \label{eq:request-representation}
\end{equation}
where $j$ denotes the position of the last non-padding token, $L$ denotes the sequence length, and $d$ is the hidden dimension.

A preference head $G_{\boldsymbol{\phi}}$ maps $\mathbf{h}$ to a vector of preference scores over the candidate models:
\begin{equation}
  \hat{\mathbf{u}}
  =
  G_{\boldsymbol{\phi}}(\mathbf{h})
  =
  (\hat{u}_1,\ldots,\hat{u}_K),
  \label{eq:router-preference}
\end{equation}
where $\hat{u}_i$ denotes the predicted preference score of candidate model $M_i$.
The resulting fixed-dimensional score vector serves as the System One decision output and can be consumed directly by the subsequent model-selection rule without autoregressive decoding or response parsing.

\textbf{Latency-Aware Model Selection.}
Although the predicted preference scores characterize the relative preferences among candidate models for the current editing request, they do not account for differences in generation latency across models. To incorporate generation efficiency into the routing decision, we use the mean end-to-end generation latency $t_i$ of each candidate model $M_i$ and convert it into a normalized speed score:
\begin{equation}
  s_i
  =
  \frac{t_{\max}-t_i}
       {t_{\max}-t_{\min}},
  \label{eq:speed-score}
\end{equation}
where $t_{\max}$ and $t_{\min}$ denote the maximum and minimum mean generation latencies among the candidate models, respectively. Consequently, faster models receive larger speed scores, with $s_i\in[0,1]$, where larger values indicate lower generation latency.

Finally, EditJev combines the preference score with the speed score:
\begin{equation}
  \hat{i}
  =
  \mathop{\arg\max}\limits_{i\in[K]}
  \left(\hat{u}_i+\beta s_i\right),
  \qquad
  \hat{M}=M_{\hat{i}},
  \label{eq:model-selection}
\end{equation}
where $\beta$ controls the quality--latency trade-off and allows training-free adjustment. Only the selected model $\hat{M}$ is executed to produce the final edited image.

\subsection{Preference-based Router Training}

\textbf{Overview.}
To train the editing router, we construct pairwise preference supervision from candidate model outputs to capture the relative suitability of different editors. For each training request, all candidate editing models are executed offline and evaluated by an MLLM-based evaluator along complementary quality dimensions. We then convert the resulting quality scores into preference pairs and optimize the router with a pairwise ranking objective.

\textbf{Pairwise Preference Construction.}
For each training request $(\mathbf{x},\mathbf{c})$, we execute every candidate model $M_i\in\mathcal{M}$ to
obtain its edited image $\mathbf{y}_i=E_i(\mathbf{x},\mathbf{c})$, where $E_i$ denotes the editing function implemented
by $M_i$. Following VIEScore~\citep{ku2024viescore}, we use an MLLM evaluator to assess editing success, preservation of content that should remain unchanged, image naturalness, and freedom from visual artifacts, each on a scale from 0 to 10. The semantic-consistency score $SC_i$ is the minimum of the editing-success and preservation scores, and the perceptual-quality score $PQ_i$ is the minimum of the naturalness and artifact-free quality scores. For this request, we define the aggregate quality score of candidate model $M_i$ as their geometric mean:
\begin{equation}
  O_i = \sqrt{SC_i\cdot PQ_i}.
  \label{eq:aggregate-quality}
\end{equation}

Given the aggregate quality scores $\mathbf{O}=(O_1,\ldots,O_K)$, we retain ordered model pairs whose quality-score difference is at least $\delta$:
\begin{equation}
  \mathcal{P}(\mathbf{x},\mathbf{c})=\{(a,b)\in[K]^2: O_a>O_b,\; O_a-O_b\geq\delta\},
  \label{eq:preference-set}
\end{equation}
 For each pair $(a,b)\in\mathcal{P}(\mathbf{x},\mathbf{c})$, candidate model $M_a$ is preferred to candidate model $M_b$. The score-gap threshold $\delta$ excludes near-tied pairs whose relative ordering may be unreliable.

\textbf{Pairwise Ranking Objective.}
Given the router prediction $\hat{\mathbf{u}}=G_{\boldsymbol{\phi}}(\mathbf{h})=(\hat{u}_1,\ldots,\hat{u}_K)$ for request $(\mathbf{x},\mathbf{c})$, we define the pairwise ranking loss as
\begin{equation}
  \ell_{\mathrm{rank}}(\mathbf{x},\mathbf{c})=
  \frac{\sum_{(a,b)\in\mathcal{P}(\mathbf{x},\mathbf{c})}
  \operatorname{softplus}(\hat{u}_b-\hat{u}_a)}
  {\max(1,|\mathcal{P}(\mathbf{x},\mathbf{c})|)},
  \label{eq:request-ranking-loss}
\end{equation}
where the sum is defined as zero when $\mathcal{P}(\mathbf{x},\mathbf{c})=\varnothing$. Given a training minibatch $\mathcal{B}$,
the overall objective is
\begin{equation}
  \mathcal{L}_{\mathrm{rank}}
  =
  \frac{1}{|\mathcal{B}|}
  \sum_{(\mathbf{x},\mathbf{c})\in\mathcal{B}}
  \ell_{\mathrm{rank}}(\mathbf{x},\mathbf{c}).
  \label{eq:ranking-loss}
\end{equation}

Compared with MCQ-style supervision, our objective avoids compressing the supervision signal into a single best-model label and can exploit multiple reliable pairwise comparisons from one request. It is also more robust than absolute score prediction, as it only relies on reliable relative preferences rather than requiring the router to match the calibrated scale of evaluator scores. Further discussion is provided in Section~\ref{sec:ablation-study}.

\begin{table*}[t]
\centering
\setlength{\tabcolsep}{2.5pt}
\renewcommand{\arraystretch}{1.2}
\caption{Quantitative comparison on ImgEdit. Methods are grouped into fixed models, MLLM-based routing methods, and System One routing methods. The best and second-best results for each editing category, overall quality, and latency are marked in bold and underline, respectively. Latency denotes the additional routing overhead before executing the selected editor.}
\resizebox{\textwidth}{!}{%
\begin{tabular}{l|ccccccccc|c|c}
\toprule
\textbf{Method} & \textbf{Add} & \textbf{Adjust} & \textbf{Extract} & 
\textbf{Replace} & \textbf{Remove} & \textbf{Background} & 
\textbf{Style} & \textbf{Compose} & \textbf{Action} & 
\textbf{Overall}$\uparrow$ & \textbf{Latency(s)}$\downarrow$\\
\midrule

\rowcolor{fixedmodel!20}
\multicolumn{12}{c}{\textbf{\emph{Fixed Models}}} \\

FLUX.2-Klein-4B 
& 4.470 & 4.660 & 1.690 & 4.480 & 4.300 & 4.470 & 4.850 & 3.510 & \underline{4.800} & 4.073 & -- \\

FLUX.2-Klein-9B 
& \underline{4.580} & \underline{4.730} & 1.950 & 4.680 & \underline{4.710} & \textbf{4.680} & \textbf{4.910} & 3.590 & \textbf{4.880} & 4.252 & -- \\

FLUX.1-Kontext-dev 
& 4.050 & 3.930 & 1.950 & 4.440 & 3.420 & 3.940 & 4.400 & 2.460 & 4.240 & 3.670 & -- \\

OmniGen2 
& 3.730 & 3.640 & 1.640 & 4.040 & 3.110 & 3.680 & 4.580 & 2.670 & 4.420 & 3.461 & -- \\

Qwen-Image-Edit 
& \textbf{4.660} & 4.650 & \textbf{4.120} & \underline{4.750} & 4.630 & 4.410 & 4.770 & \textbf{3.770} & 4.750 & \underline{4.540} & -- \\

\midrule

\rowcolor{mllmrouting!20}
\multicolumn{12}{c}{\textbf{\emph{MLLM Routing}}} \\

GPT-5.5 Routing 
& 4.255 & 4.496 & 3.994 & 4.735 & 4.577 & 4.304 & 4.674 & 3.072 & 4.509 & 4.380 & 11.85 \\

Gemini-3.7-Flash Routing 
& 4.048 & 4.287 & 3.906 & 4.441 & 3.740 & 3.964 & 4.507 & 3.203 & 4.639 & 4.123 & 17.68 \\

GLM-5.3-Flash Routing 
& 4.132 & 4.060 & 3.242 & 4.634 & 4.154 & 4.297 & 4.753 & 2.898 & 4.602 & 4.137 & 4.39 \\

Kimi-K3 Routing 
& 4.385 & 4.578 & 2.949 & 4.554 & 3.980 & 4.130 & 4.750 & 2.956 & 4.509 & 4.134 & 14.50 \\

\midrule

\rowcolor{systemone!20}
\multicolumn{12}{c}{\textbf{\emph{System One Routing}}} \\

Jev Routing 
& 4.349 & \underline{4.730} & 2.262 & 4.543 & 4.457 & 4.141 & 4.830 & 3.405 & \textbf{4.880} & 4.138 & \underline{0.87} \\

\textbf{EditJev} 
& 4.543 & \textbf{4.762} & \underline{4.088} & \textbf{4.772} & \textbf{4.716} & \underline{4.662} & \underline{4.890} & \underline{3.651} & 4.741 & \textbf{4.592} & \textbf{0.094} \\

\bottomrule
\end{tabular}%
}

\label{tab:imgedit}
\end{table*}

\section{Experiments}

\subsection{Experimental Setup}

\textbf{Implementation Details.}
We use Qwen3-VL-4B-Instruct~\citep{bai2025qwen3} as the multimodal router backbone, freezing its pretrained parameters and optimizing LoRA adapters and a linear preference head. The candidate pool consists of FLUX.2 Klein 4B, FLUX.2 Klein 9B, FLUX.1-Kontext-dev, OmniGen2, and Qwen-Image-Edit. 
$\beta$ is set as $0.7$ and $0.02$ on ImgEdit and GEdit-En. All latency measurements are conducted on an NVIDIA RTX PRO 6000 GPU using each model's official inference configuration. We measure the generation time of each model over 20 different test requests and report the average.  We set the pairwise score-gap threshold to $\delta=0.5$. The routing latency of each routing method is measured separately as the additional overhead before executing the selected editor. For closed-source routing models, including API-based MLLM routers and Jev, the latency is measured as the API response time and averaged over 5 requests. For EditJev, the latency is measured using local inference and averaged over 5 requests.
 
For ablations of the training objective, visual input, and backbone scale, we fix $\beta=0$ to evaluate their effects on routing quality. We separately vary $\beta$ to study the inference-time quality--latency trade-off.
More details are provided in the appendix.

\textbf{Compared Methods.}
We compare EditJev with fixed-model policies, general-purpose MLLM routing baselines, and System-One routing baselines. Fixed-model policies always invoke the same candidate model; we report each candidate separately. For general-purpose MLLM routing, we evaluate GPT-5.5~\citep{openai2026gpt55}, Gemini-3.7-Flash~\citep{google2026gemini37flash}, GLM-5.3-Flash~\citep{zai2026glm53flash}, and Kimi-K3~\citep{moonshot2026k3} on the same 737 ImgEdit and 606 GEdit-En requests. Each MLLM receives the source image, editing instruction, candidate model names, and their measured mean generation times through the same prompt template. It selects the editor expected to provide the highest quality, preferring faster editors when the expected quality is similar. Candidate outputs and reference scores are not included in the prompt. Jev is included as a System-One routing baseline that directly predicts preferences over candidate models without autoregressive model selection.

\textbf{Benchmarks.}
We evaluate EditJev on the ImgEdit~\citep{ye2026imgedit} and GEdit-En~\citep{liu2025step1x} image-editing benchmarks, using GPT-5.1 as the primary evaluator and following the official evaluation protocols. We report the benchmark-defined metrics for each dataset.

\begin{figure*}[t]
\centering
\includegraphics[width=\textwidth]{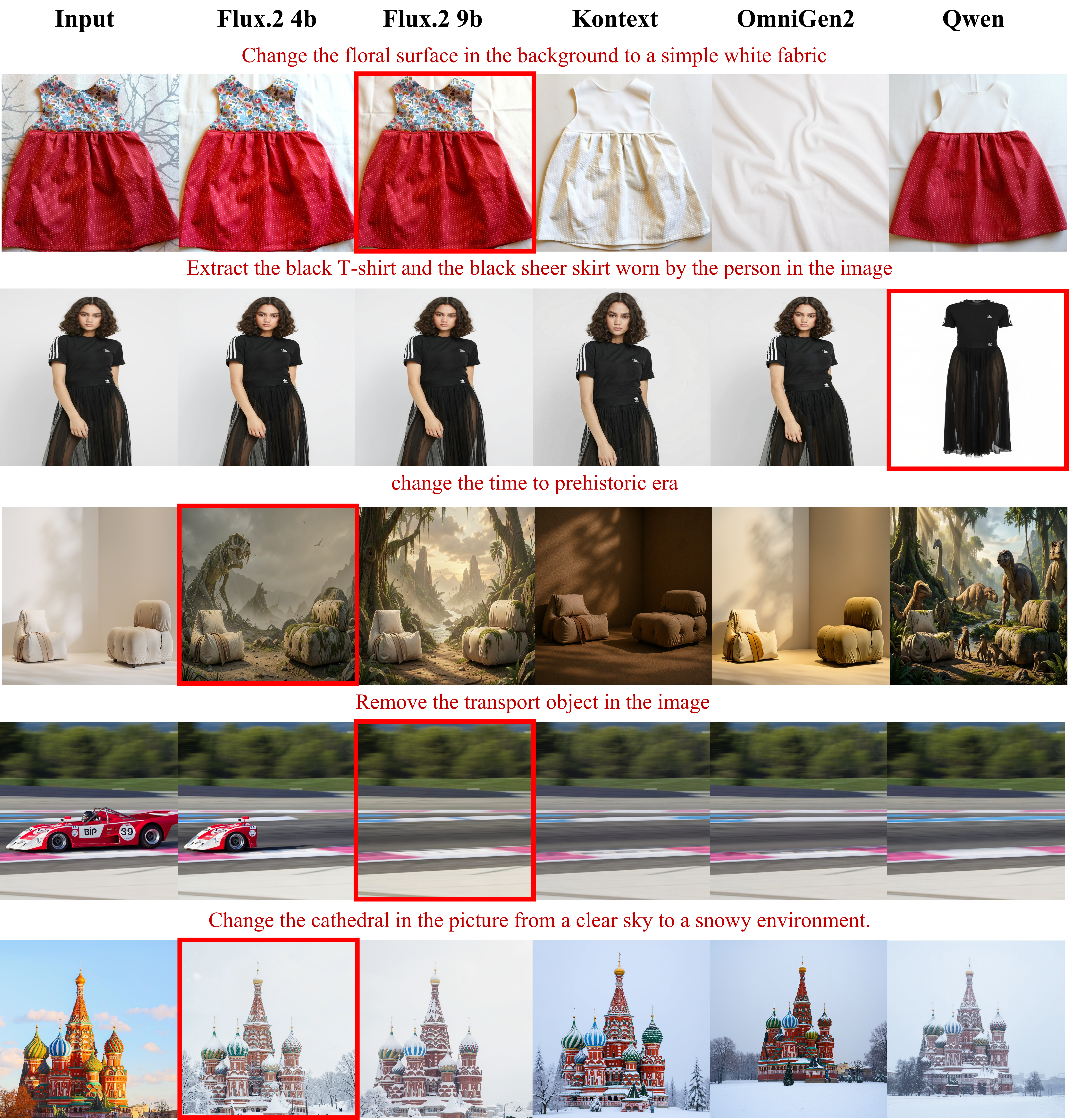}
\caption{Qualitative routing results on GEdit-En and ImgEdit. A red border marks the model selected by EditJev for each input.}
\label{fig:qualitative}
\end{figure*}

\subsection{Main Results}

\begin{wraptable}{r}{0.55\textwidth}
\vspace{-20pt}
\caption{Quantitative evaluation results on GEdit-En.}
\centering
\small

\renewcommand{\arraystretch}{0.9}

\begin{tabular}{l|cc|c}
\toprule
\textbf{Method} & \textbf{G\_SC} & \textbf{G\_PQ} & \textbf{G\_O}$\uparrow$ \\
\midrule

\rowcolor{fixedmodel!20}
\multicolumn{4}{c}{\textbf{\emph{Fixed Models}}} \\

FLUX.2-Klein-4B & 7.731 & 8.397 & 7.667 \\
FLUX.2-Klein-9B & 8.081 & \textbf{8.416} & 7.965 \\
FLUX.1 Kontext & 6.404 & 7.579 & 6.241 \\
OmniGen2 & 6.391 & 7.762 & 6.297 \\
Qwen-Image-Edit & \underline{8.441} & 8.326 & \underline{8.194} \\

\midrule

\rowcolor{mllmrouting!20}
\multicolumn{4}{c}{\textbf{\emph{MLLM Routing}}} \\

GPT-5.5 Routing & 8.238 & 8.110 & 7.920 \\
Gemini-3.7-Flash Routing & 7.244 & 7.732 & 6.996 \\
GLM-5.3-Flash Routing & 7.652 & 7.910 & 7.402 \\
Kimi-K3 Routing & 7.760 & 8.035 & 7.543 \\

\midrule

\rowcolor{systemone!20}
\multicolumn{4}{c}{\textbf{\emph{System One Routing}}} \\

Jev Routing & 7.913 & 8.281 & 7.762 \\
\textbf{EditJev} & \textbf{8.452} & \underline{8.409} & \textbf{8.255} \\

\bottomrule
\end{tabular}

\vspace{-10pt}
\label{tab:gedit}
\end{wraptable}

\textbf{Quantitative Comparison.}
Table~\ref{tab:imgedit} and~\ref{tab:gedit} compare EditJev with fixed-model policies and general-purpose MLLM routing baselines. On ImgEdit, EditJev achieves an overall score of 4.592, surpassing the strongest fixed-model baseline, Qwen-Image-Edit, which obtains a score of 4.540. On GEdit-En, EditJev achieves a $G_O$ score of 8.255, compared with 8.194 from Qwen-Image-Edit. These results demonstrate that sample-level routing can effectively exploit the complementary capabilities of different editing models rather than relying on a single globally preferred model. Compared with GPT-5.5 Routing, a general-purpose strong MLLM baseline, EditJev achieves higher aggregate scores on both benchmarks, improving from 4.380 to 4.592 on ImgEdit and from 7.920 to 8.255 on GEdit-En. Moreover, following the System One design principle of direct structured decision output, both Jev and EditJev introduce lower routing overhead than general-purpose MLLM routers. While Jev requires 0.87s per request, EditJev further reduces the routing latency to 0.094s per request through a lightweight preference prediction head.

\begin{wraptable}{r}{0.5\textwidth}

\vspace{-20pt}
\caption{Latency comparison among Qwen-Image-Edit, GPT-5.5 Routing, and EditJev. Latency includes both the routing inference time and the generation time of the selected editor. }
\centering
\setlength{\tabcolsep}{2.5pt}
\begin{tabular}{l|cc}
\toprule
\textbf{Method} & \textbf{ImgEdit (s)}$\downarrow$ & \textbf{GEdit-En (s)}$\downarrow$  \\
\midrule
Qwen-Image-Edit & 56.90 & 56.90 \\
GPT-5.5 Routing & \underline{51.15} & \underline{53.55} \\
\textbf{EditJev} & \textbf{22.33} & \textbf{38.88} \\
\bottomrule
\end{tabular}

\vspace{-10pt}
\label{tab:latency}
\end{wraptable}

\textbf{Efficiency Comparison.} As shown in Table~\ref{tab:latency}, we compare the end-to-end latency of EditJev with Qwen-Image-Edit and GPT-5.5 Routing, which represent strong fixed-model deployment and general-purpose MLLM-based routing strategies, respectively. The reported latency includes both routing inference and selected-model generation time. Compared with Qwen-Image-Edit, the fixed model with competitive average quality, EditJev reduces the end-to-end latency from 56.9 to 22.3 seconds on ImgEdit and from 56.9 to 38.9 seconds on GEdit-En, achieving $2.55\times$ and $1.46\times$ inference speedups, respectively, while improving the aggregate scores on both benchmarks. This efficiency gain is enabled by EditJev's ability to select faster candidates when their predicted preference scores are comparable, while still selecting slower models when their predicted quality advantages justify the additional latency.

GPT-5.5 Routing achieves end-to-end latencies of 51.15 and 53.55 seconds on ImgEdit and GEdit-En, respectively, which are higher than EditJev's latency. Meanwhile, GPT-5.5 Routing obtains lower aggregate quality scores than both Qwen-Image-Edit and EditJev on both benchmarks (Table~\ref{tab:imgedit} and Table~\ref{tab:gedit}). Overall, EditJev achieves a better quality--latency trade-off compared with both fixed-model deployment and general-purpose MLLM-based routing strategies.

\textbf{Qualitative Comparison.}
Fig.~\ref{fig:qualitative} illustrates how EditJev adapts its selection to editing quality and model latency. For the instruction \emph{``Change the floral surface in the background to a simple white fabric''} , only FLUX.2 Klein 4B and FLUX.2 Klein 9B correctly replace the patterned background with simple white fabric, whereas Qwen-Image-Edit, despite being the strongest fixed model on average, incorrectly alters the garment itself. Among the two successful candidates, FLUX.2 Klein 9B better preserves the original garment colors than FLUX.2 Klein 4B, while incurring only a small increase in generation latency, leading EditJev to select it.
For the instruction \emph{``Extract the black T-shirt and the black sheer skirt worn by the person in the image,''} only Qwen-Image-Edit successfully extracts the requested garments, and EditJev selects this model despite its higher generation latency.
For the instruction \emph{``Remove the transport object in the image''}, all candidates except FLUX.2 Klein 4B successfully remove the racing car. Among these successful candidates, FLUX.2 Klein 9B has the lowest mean generation time and is therefore selected. These examples show that EditJev does not simply choose the strongest or fastest model globally; instead, it selects a model whose output quality is suitable for the specific request while accounting for generation efficiency. This adaptive behavior highlights the advantage of sample-level routing, where different editors can be utilized according to their strengths across diverse editing scenarios.

\subsection{Ablation Study}

\begin{wraptable}{r}{0.4\textwidth}
\vspace{-20pt}
\caption{Ablation on the router training objective on the ImgEdit benchmark.}
\centering
\small
\setlength{\tabcolsep}{3pt}
\begin{tabular}{lc}
\toprule
\textbf{Training Objective} & \textbf{Overall} $\uparrow$ \\
\midrule
MSE Regression & 4.5545 \\
MCQ Generation & 4.5491 \\
Pairwise Ranking (Ours) & \textbf{4.6002} \\
\bottomrule
\end{tabular}

\label{tab:ablation-objective}
\end{wraptable}

\label{sec:ablation-study}

\textbf{Effect of the Ranking Objective.}
We compare the pairwise ranking objective with two alternative training formulations. 
\textbf{MSE Regression} replaces the pairwise ranking loss with mean squared error regression on the candidate-quality scores, requiring the router to predict absolute evaluator scores. 
\textbf{MCQ Generation} formulates routing as multiple-choice generation, where the router is trained to output only the index of the highest-scoring candidate model.
As shown in Table~\ref{tab:ablation-objective}, MSE regression reduces the overall score on ImgEdit from 4.6002 to 4.5545, while MCQ generation further decreases it to 4.5491. These results indicate that pairwise preference learning is more effective for routing than either absolute score regression or single-label generation. The ranking objective directly optimizes the relative ordering among candidate models and can exploit multiple reliable pairwise comparisons from each request. In contrast, MCQ generation compresses the supervision signal into a single best-model label, discarding the preference relationships among the remaining candidates. MSE regression requires the router to accurately predict the evaluator's absolute scores, making the learning problem more challenging.

\begin{wraptable}{r}{0.4\textwidth}
\vspace{-20pt}
\caption{Ablation on the visual input to the router on ImgEdit.}
\centering
\small
\setlength{\tabcolsep}{3pt}
\begin{tabular}{lc}
\toprule
\textbf{Variant} & \textbf{Overall} $\uparrow$ \\
\midrule
w/o Image Input & 4.5717 \\
w/ Image Input (Ours) & \textbf{4.6002} \\
\bottomrule
\end{tabular}

\label{tab:ablation-image}
\end{wraptable}

\textbf{Effect of Multimodal Input.} Shown in Table~\ref{tab:ablation-image}, w/o Image Input removes the source image from the router input and performs model selection solely based on the editing instruction. Removing the source image reduces the overall score on ImgEdit from 4.6002 to 4.5717. This result indicates that, even for the same editing instruction, the suitability of different candidate models may vary depending on the source image. Therefore, the router should consider both the editing instruction and the visual content of the source image. Unlike many LLM and text-to-image routing settings that primarily rely on textual inputs, model routing for image editing should jointly consider the editing instruction and the source image.

\begin{wraptable}{r}{0.4\textwidth}

\centering
\small
\setlength{\tabcolsep}{3pt}
\vspace{-10pt}
\caption{Ablation on the router backbone scale on ImgEdit.}
\begin{tabular}{lc}
\toprule
\textbf{Router Backbone} & \textbf{Overall} $\uparrow$ \\
\midrule
Qwen3-VL-2B & 4.5898 \\
Qwen3-VL-4B (Ours) & \textbf{4.6002} \\
Qwen3-VL-32B & 4.5898 \\
\bottomrule
\end{tabular}

\label{tab:ablation-scale}

\end{wraptable}

\textbf{Effect of Router Backbone Scale.}
We further investigate the impact of different router backbone sizes. As shown in Table~\ref{tab:ablation-scale}, EditJev achieves comparable routing performance across different Qwen3-VL backbone scales. The results show that increasing the backbone size does not consistently improve routing performance. These results suggest that EditJev is not dependent on a specific backbone scale and can maintain stable routing performance across different multimodal backbones.

\begin{wrapfigure}{r}{0.4\textwidth}
\vspace{-20pt}
\centering
\includegraphics[width=\linewidth]{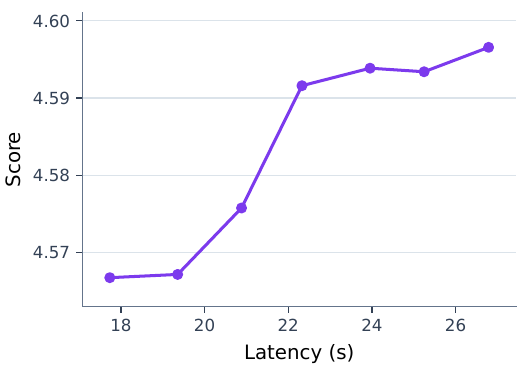}
\caption{Quality--latency trade-off on ImgEdit under different speed weights.}
\label{fig:imgedit-speed-tradeoff}
\vspace{-20pt}
\end{wrapfigure}

\textbf{Effect of the Speed Weight.}
We trace the quality--latency trade-off on ImgEdit by varying the inference-time speed weight $\beta$ from $0.4$ to $1.0$ in increments of $0.1$, while keeping the trained router fixed (Fig.~\ref{fig:imgedit-speed-tradeoff}). As $\beta$ increases, estimated average end-to-end latency—computed by averaging the selected models' measured mean generation times and adding router overhead—falls from 26.79 to 17.73 seconds (33.8\%), while the overall score changes from 4.5966 to 4.5667. The main setting, $\beta=0.7$, achieves a score of 4.5916 at 22.33 seconds, only 0.0050 below the score at $\beta=0.4$. Thus, the speed score allows inference-time adjustment of the quality--latency trade-off without retraining the router.

\begin{wraptable}{r}{0.5\textwidth}

\centering
\small
\setlength{\tabcolsep}{3pt}
\vspace{-20pt}
\caption{Human preference alignment on 20 cases from ImgEdit and GEdit-En.}
\begin{tabular}{lcc}
\toprule
\textbf{Method} & \textbf{Agreement (\%)} $\uparrow$ & \textbf{Spearman} $\rho$ $\uparrow$ \\
\midrule
GPT-5.5 Routing & 34.50 & --- \\
EditJev (Ours) & \textbf{81.75} & 0.7478 \\
\bottomrule
\end{tabular}

\label{tab:user_study}
\end{wraptable}

\subsection{User Study}
We assess human preference alignment on 20 cases from ImgEdit and GEdit-En. For each case, 20 participants viewed the source image, editing instruction, and outputs of five candidate models alongside the measured mean generation times of the candidate models, then selected a preferred result based on quality and efficiency (400 judgments total). The model selected by EditJev was identical to the participant-selected model in 327 out of 400 individual judgments (81.75\%), compared with 138 (34.50\%) for GPT-5.5 routing—a 47.25 percentage-point difference (Table~\ref{tab:user_study}). Agreement rates are computed over all 400 judgments. We also computed, for each case, the Spearman correlation between candidate vote counts and EditJev's final speed-adjusted routing scores (Eq.~\ref{eq:model-selection}), including zero-vote candidates and assigning average ranks to ties. The mean correlation across cases was $\rho=\mathbf{0.7478}$, indicating positive alignment between EditJev's ranking and human preferences. We cannot compute a comparable correlation for GPT-5.5 because its recorded responses include only the selected model, not scores or a full ranking.

\section{Conclusion}

To the best of our knowledge, this paper is the first to formulate sample-level image-editing model routing, where a router dynamically selects a suitable editing model for each source image and editing instruction. We introduce EditJev, a System One routing framework that learns preferences over candidate editing models and selects suitable editors for different image--instruction pairs, enabling quality--latency-aware and efficient routing.

Future work will explore more reliable and scalable routing supervision. In particular, we plan to investigate stronger multimodal evaluation frameworks, including agentic multimodal judges that can provide more comprehensive and fine-grained assessments of editing quality, to obtain more reliable preference signals. We also aim to extend routing to larger and dynamically evolving pools of editing models, developing more robust routers that support the addition and removal of editors without extensive retraining. These directions may further improve the scalability and adaptability of image editing model routing systems.

\bibliography{iclr2027_conference}
\bibliographystyle{iclr2027_conference}

\end{document}